%% file: example_paper.tex
\documentclass[accepted]{article}

\usepackage{microtype}
\usepackage{graphicx}
\usepackage{subfigure}
\usepackage{booktabs} 

\usepackage{hyperref}

\usepackage{icml2025}

\usepackage{amsmath}
\usepackage{amssymb}
\usepackage{mathtools}
\usepackage{amsthm}
\usepackage{makecell}

\usepackage{amsfonts}
\usepackage{multirow}
\usepackage{ulem}

\newcommand{\purple}[1]{{\color{purple} #1}}
\def\sysname{Lumina-Video}

\usepackage[capitalize,noabbrev]{cleveref}

\theoremstyle{plain}

\theoremstyle{definition}

\theoremstyle{remark}

\usepackage[textsize=tiny]{todonotes}

\icmltitlerunning{Lumina-Video: Efficient and Flexible Video Generation with Multi-scale Next-DiT}

\begin{document}

\twocolumn[
\icmltitle{Lumina-Video: Efficient and Flexible Video Generation \\ with Multi-scale Next-DiT}

\icmlsetsymbol{equal}{$\textbf{*}$} 
\icmlsetsymbol{corresp}{$\P$} 
\icmlsetsymbol{lead}{$\blacklozenge$} 
\icmlsetsymbol{sep}{\!\!,\!} 

\begin{icmlauthorlist}
\icmlauthor{Dongyang Liu}{equal,CUHK,AILAB}
\icmlauthor{Shicheng Li}{equal,AILAB}
\icmlauthor{Yutong Liu}{equal,AILAB}
\icmlauthor{Zhen Li}{equal,CUHK}
\icmlauthor{Kai Wang}{equal,AILAB}
\icmlauthor{Xinyue Li}{equal,AILAB}
\icmlauthor{Qi Qin}{AILAB}
\icmlauthor{Yufei Liu}{AILAB}
\icmlauthor{Yi Xin}{AILAB}
\icmlauthor{Zhongyu Li}{AILAB,NKU}
\icmlauthor{Bin Fu}{AILAB}
\icmlauthor{Chenyang Si}{AILAB}
\icmlauthor{Yuewen Cao}{AILAB}
\icmlauthor{Conghui He}{AILAB}
\icmlauthor{Ziwei Liu}{AILAB}
\icmlauthor{Yu Qiao}{AILAB}
\icmlauthor{Qibin Hou}{NKU}
\icmlauthor{Hongsheng Li}{corresp,CUHK,AILAB}
\icmlauthor{Peng Gao}{corresp,lead,AILAB}
\end{icmlauthorlist}


\icmlaffiliation{CUHK}{The Chinese University of Hong Kong}
\icmlaffiliation{AILAB}{Shanghai AI Laboratory}
\icmlaffiliation{NKU}{Nankai University}

\icmlcorrespondingauthor{Peng Gao}{gaopeng@pjlab.org.cn}
\icmlcorrespondingauthor{Hongsheng Li}{hsli@ee.cuhk.edu.hk}


\icmlkeywords{Machine Learning, ICML}

\vskip 0.3in
]
    



\printAffiliationsAndNotice{%
    \textsuperscript{\textbf{*}}Equal Contribution
    \textsuperscript{$\P$}Corresponding Authors
    \textsuperscript{$\blacklozenge$}Project Lead}

\input{section/0_abstract}
\input{section/1_intro}
\input{section/2_related_work}
\input{section/3_methods}
\input{section/4_experiments}
\input{section/5_conclusion}

\nocite{langley00}

\bibliography{example_paper}
\bibliographystyle{icml2025}

\newpage
\appendix
\onecolumn
\input{section/supp}

\end{document}

%% file: section/0_abstract.tex
\begin{abstract}

Recent advancements have established Diffusion Transformers (DiTs) as a dominant framework in generative modeling. Building on this success, Lumina-Next achieves exceptional performance in the generation of photorealistic images with Next-DiT. However, its potential for video generation remains largely untapped, with significant challenges in modeling the spatiotemporal complexity inherent to video data. To address this, we introduce Lumina-Video, a framework that leverages the strengths of Next-DiT while introducing tailored solutions for video synthesis. 
Lumina-Video incorporates a Multi-scale Next-DiT architecture, which jointly learns multiple patchifications to enhance both efficiency and flexibility.
By incorporating the motion score as an explicit condition,~\sysname{} also enables direct control of generated videos' dynamic degree. 
Combined with a progressive training scheme with increasingly higher resolution and FPS, and a multi-source training scheme with mixed natural and synthetic data,~\sysname{} achieves remarkable aesthetic quality and motion smoothness at high training and inference efficiency. We additionally propose Lumina-V2A, a video-to-audio model based on Next-DiT, to create synchronized sounds for generated videos. Codes are released at \url{https://www.github.com/Alpha-VLLM/Lumina-Video}.

\end{abstract}

%% file: section/1_intro.tex
\section{Introduction}
\label{sec:intro}

The field of generative modeling has witnessed significant advancements in recent years, with Diffusion Transformers (DiTs) emerging as a powerful paradigm for creating high-quality photorealistic content~\citep{DiT,sd3,openai2024sora}. 
One notable innovation is Next-DiT~\citep{luminanext}, an improved version of flow-based DiT that has shown strong performance in image generation.
By combining architectural enhancements such as 3D RoPE for superior spatiotemporal representation, sandwich normalization for stabilized training, and grouped-query attention for efficient attention computation, Next-DiT has achieved remarkable success. The model consistently produces images that are not only visually compelling but also exhibit high diversity and fine-grained details. 
However, despite its advancements in image synthesis, the potential of Next-DiT for video generation remains underexplored. 

Video generation poses unique challenges that go beyond those encountered in image generation. The inherent complexity of modeling both spatial and temporal dimensions in a coherent manner introduces significant computational and architectural challenges. While Next-DiT can be adapted for video tasks, its current design is not specifically tailored for the spatiotemporal intricacies of video data, leading to an excessive number of video tokens and low computational efficiency. 
These limitations underscore the need for a tailored approach that fully leverages the capabilities of Next-DiT while addressing the unique demands of video synthesis.

\begin{figure*}[t]
    \centering
    \includegraphics[width=1\linewidth]{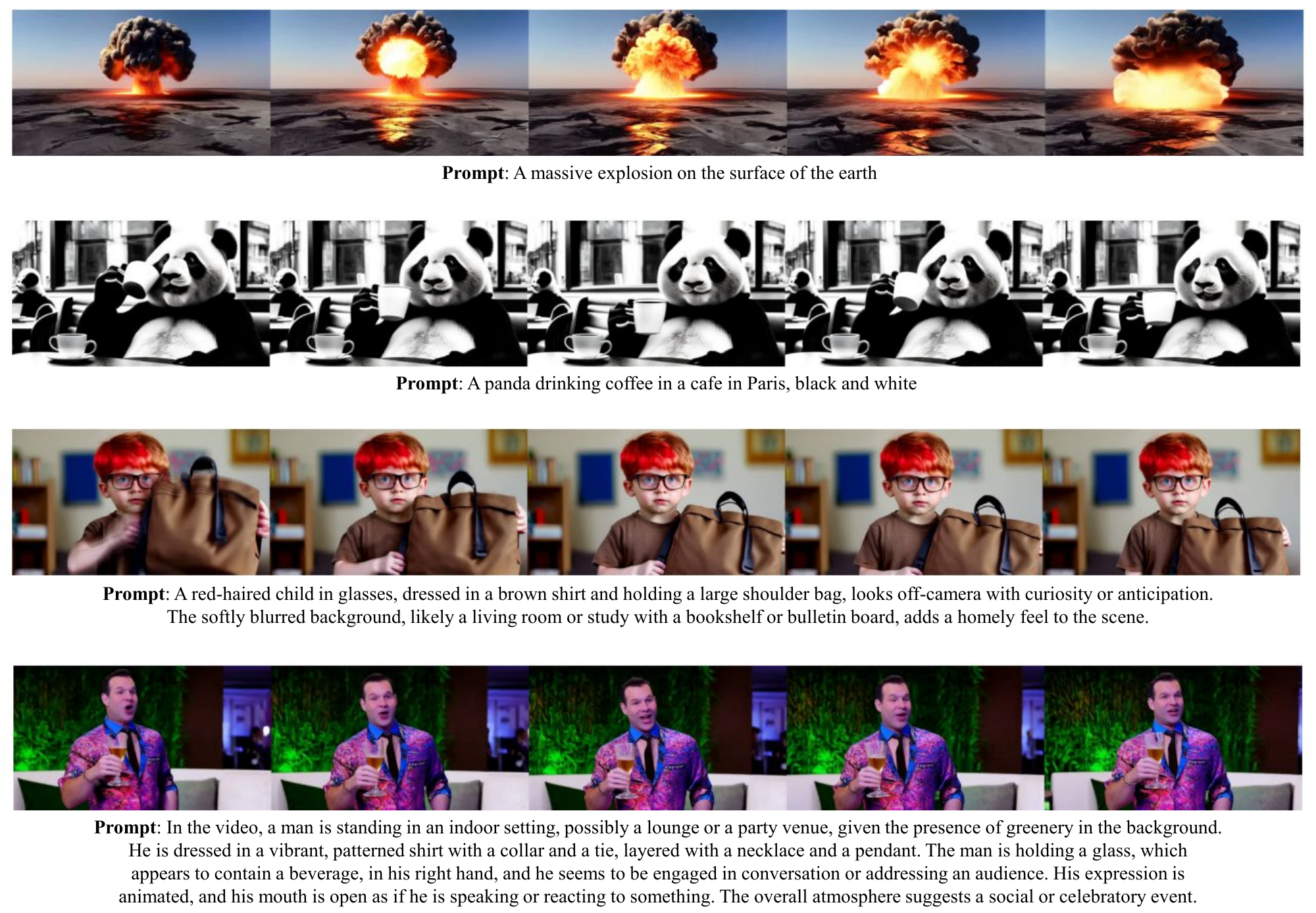}
    \vspace{-1.5em}
    \caption{Lumina-Video demonstrates a strong ability to generate high-quality videos with rich details and remarkable temporal coherence, accurately following both simple and detailed text prompts.}
    \label{fig:demos}
\end{figure*}

To bridge this gap, we introduce Lumina-Video, a novel framework that excels at generating high-quality videos by building upon the strengths of Next-DiT. 
At the core of Lumina-Video is Multi-scale Next-DiT, an extension of Next-DiT into a multi-scale architecture by introducing multiple patch sizes that share a common DiT backbone and are trained jointly in a unified manner. 
This straightforward yet elegant approach allows the model to learn video structures across different computational budgets simultaneously. By strategically allocating different patchifications to various sampling timesteps,~\sysname{} achieves notable improvements in inference efficiency with only a minor sacrifice in quality. This design also enables users to dynamically adjust the computational cost based on resource constraints and specific requirements, offering greater flexibility during inference.
Considering the importance of motion in videos, we additionally derive motion scores from optical flow and incorporate them as an extra conditioning input to the DiT. By designing a systematic strategy that separately manipulates the motion conditioning of positive and negative classifier-free guidance (CFG)~\citep{cfg} samples,~\sysname{} provides an effective interface for controlling the extent of dynamics in generated videos. 
We further refine the training strategy by progressively training the model on videos with increasing spatiotemporal resolutions to improve training efficiency, leveraging joint image-video training to enhance frame quality and text comprehension, and incorporating multi-source training to fully utilize diverse real and synthetic data sources. These designs enable Lumina-Video to seamlessly tackle the challenging video generation task across a wide range of scenarios.

Our contributions establish Lumina-Video as a new solution for video generation, 
offering researchers and practitioners a powerful and flexible tool for creating video content. 
By fully unleashing the potential of Multi-Scale Next-DiT, Lumina-Video is capable of generating high-fidelity videos of varying resolution, excelling in both quantitative metrics and visual quality. 
In addition, we design a Lumina-V2A, a Next-DiT-based video-to-audio framework to bring generated silent videos to real life with synchronized sounds. 
Furthermore, in line with our commitment to democratizing access to advanced video generation technologies, we open-source our training framework and model parameters, empowering the research community to explore, extend, and deploy Lumina-Video across a diverse range of applications. 
Through this work, we aim to catalyze future innovations in video generation and pave the way for broader adoption of generative modeling techniques.

%% file: section/2_related_work.tex
\section{Related Work}
\label{sec:related_work}
\subsection{Video Generation}

The field of video generation has evolved rapidly, combining advances in image synthesis with the additional complexity of temporal consistency across video frames. 
Early models predominantly relied on GANs for video generation~\citep{TGAN,StyleGAN-V,MoCoGAN}, which were capable of producing videos with rich details but often suffered from unstable training and mode collapse. 
More recently, following breakthroughs in image generation and language modeling, the landscape of video generation has expanded to include diverse paradigms such as masked modeling~\citep{Phenaki,MAGVIT,MAGVIT-v2}, autoregressive modeling~\citep{CogVideo,VideoPoet,Video-LaViT,luminamgpt}, and diffusion models~\citep{PYoCo,WALT,StableVideoDiffusion,CogVideoX}. 
Among these, diffusion models have emerged as the dominant approach due to their ability to produce high-quality videos with exceptional temporal consistency, as evidenced by their adoption in state-of-the-art proprietary systems like Sora~\citep{openai2024sora}. However, the high computational cost of training diffusion-based text-to-video models remains a significant obstacle. 
In this work, we address this challenge by introducing Lumina-Video, a multi-scale diffusion-based framework, achieving remarkable results in text-to-video generation with reduced computational burden.

\subsection{Transformer-based Diffusion Models}

Early diffusion models~\citep{DDPM,DDIM} primarily relied on convolution-based U-Nets~\citep{u-net}. However, the rise of transformer architectures has revolutionized computer vision by achieving state-of-the-art performance across various tasks~\citep{vit,detr,mae}. Building on this success, diffusion transformers like DiT~\citep{DiT} and UViT~\citep{UViT} successfully adapted transformer architectures for visual generation, significantly advancing the field. This approach has since become the dominant paradigm in diffusion-based models, as demonstrated by the wide adoption in both open-source models and commercial systems~\citep{sd3,latte,pixartalpha,pixartsigma,openai2024sora}. Recent work has also explored flow matching as an alternative to standard diffusion processes, offering improved efficiency and flexibility~\citep{FlowMatching,FlowMatchingGuide,RectifiedFlow,SiT}. Notably, models like Lumina-T2X~\citep{luminat2x} and Lumina-Next~\citep{luminanext} have refined the core components of flow-based diffusion transformers, achieving remarkable performance in image synthesis. Our work builds on these advances, extending the framework to video generation. By incorporating multi-scale learning and motion-aware conditioning, we adapt and enhance diffusion transformers to efficiently capture the complex spatiotemporal dynamics of videos.

\subsection{Multi-scale Learning in Computer Vision}

The concept of multi-scale processing has long been fundamental in computer vision~\citep{Koenderink2004TheSO,Burt1983TheLP,Adelson1984PYRAMIDMI}.
With the advent of deep learning, the idea of multi-scale processing has been revitalized, yielding significant benefits for tasks that demand both high-level semantics and low-level details~\citep{FPN,deeplab,PredictingDepth,multiscaleinteractive,AttentionTosSale,AUnifiedMulti,multiscalehigh,multiscalevision,swintransformer,featurepyramidtransformer}. 
In the context of visual generation, recent advances have embraced multi-scale architectures to enhance generative processes~\citep{hierarchicalpatchdiffusion,xiaoyu2024multiscale,matryoshkadiffusion,lego,PyramidFlow}. 
Building on these foundational works, we introduce a novel Multi-scale Next-DiT architecture for video generation. Our method extends the multi-scale paradigm to the spatiotemporal domain, learning video structures across multiple levels of detail with varying patch sizes. This design not only ensures efficient training but also excels in generating high-quality, temporally coherent videos.

%% file: section/3_methods.tex
\begin{figure*}[t]
    \centering
    \includegraphics[width=1\linewidth]{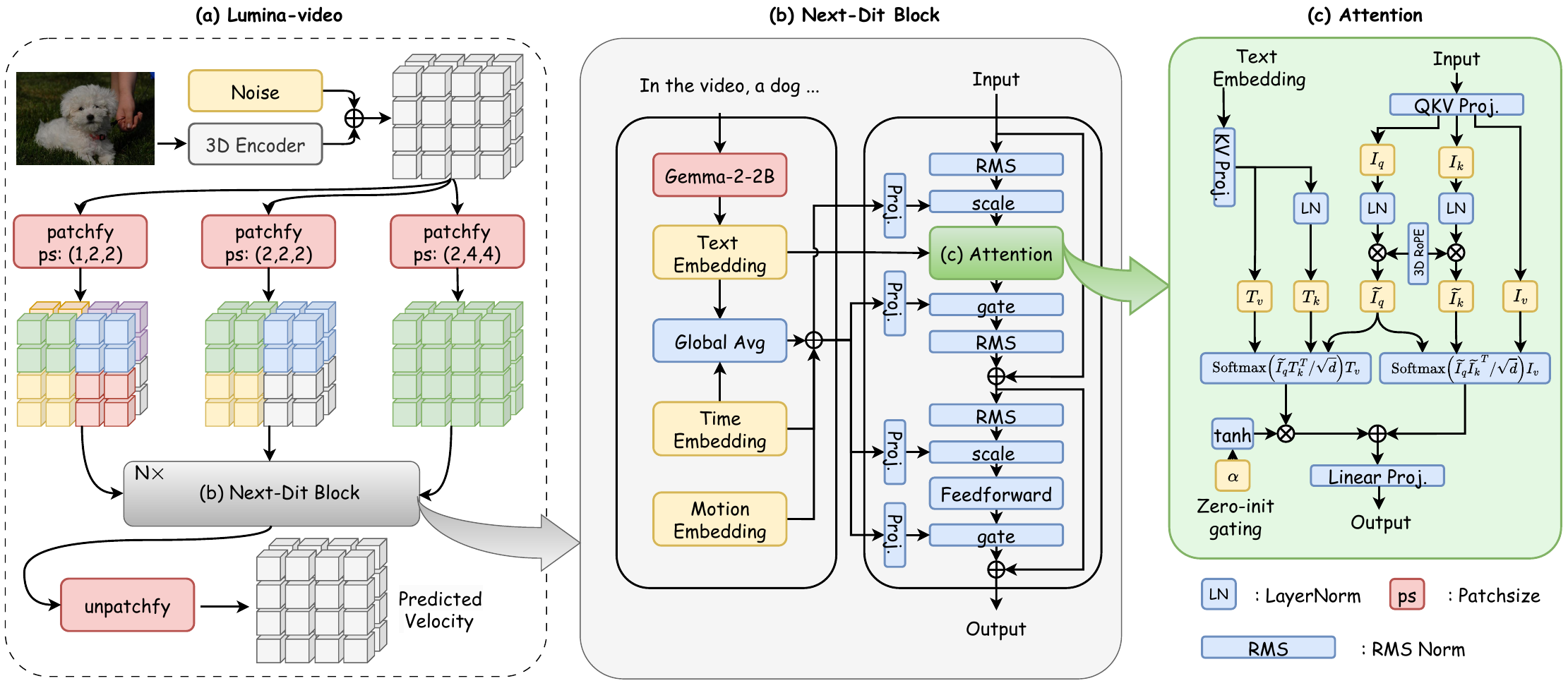}
    \vspace{-1em}
    \caption{Architecture of Lumina-Video with Multi-scale Next-DiT and Motion Conditioning.}
    \label{fig:architecture}
\end{figure*}

\section{Lumina-Video}
\label{sec:method}

In this section, we elaborate on the multi-patch method and the explicit injection of motion motioning. An introduction to the preliminaries, including the VAE, the text encoder, the Next-DiT architecture, and the loss function, is deferred to Sec.\ref{sec:basic_comp} in the appendix. A graphical illustration of the overall architecture is provided in Fig.\ref{fig:architecture}. An extra video-to-audio extension is introduced in Sec.~\ref{sec:V2A}.

\subsection{Multi-scale Next-DiT}
\label{sec:ms-next-dit}

In video generation based on diffusion transformers, the number of tokens processed by the transformer plays a pivotal role in determining the computational cost and training efficiency. More tokens allows the DiT to capture more fine-grained details, while it involves increased computational cost and degrades the efficiency

In this work, we propose Multi-scale Next-DiT, a novel architecture that incorporates multiple pairs of patchify and unpatchify layers trained in a unified manner. This architecture enables a systematic analysis of the impact of token quantities on the denoising process and demonstrates improved efficiency. By leveraging multiple scales, the model enables high efficiency by properly combining multiple scales in one complete denoising process. Moreover, our approach offers great flexibility that allows the model to adapt dynamically to diverse requirements in the inference stage. This adaptability advances the development and deployment of text-to-video models in multiple practical scenarios.

\subsubsection{Multi-scale Patchification}

In DiT-based T2V models, given a fixed VAE, the number of tokens is determined by two operations: patchify and unpatchify. 
The patchify layer converts the noised latent representation $\mathbf{z}_t \in \mathbb{R}^{T \times H \times W \times C}$ into a sequence of tokens via a linear transformation before undergoing the DiT blocks, where the number of tokens can be calculated as 
\begin{equation}
    N = \frac{T \times H \times W}{p_t \times p_h \times p_w}
\end{equation}
$(p_t, p_h, p_w)$ denotes the patch size, a critical hyperparameter controlling the granularity of the representation. 
After the DiT blocks, the output tokens are linearly projected and reshaped back into the original shape by an unpatchify layer.

To endow the model with the ability to flexibly handle different levels of granularity based on varying computational requirements, we introduce the core component of Multi-scale Next-DiT: multi-scale patchification. Instead of using a single patch size, we instantiate multiple pairs of patchify and unpatchify layers with different spatio-temporal patch sizes, each corresponding to a distinct scale. Specifically, we employ a hierarchy of $M$ patch sizes
$\{P^i=(p_t^i, p_h^i, p_w^i)|i=1,\cdots,M\}$, where $p_t^i \leq p_t^{i+1}$, $p_w^i \leq p_w^{i+1}$, $p_h^i \leq p_h^{i+1}$. 
Patchifying with a larger patch size 
leads to greater computation reduction, while patchifying with a smaller patch size preserves finer details in the latent representations. This allows us to dynamically adjust the level of abstraction used according to our demand. 

Note that in Multi-scale Next-DiT, all patch sizes share the same DiT backbone. This design minimizes parameter count and memory overhead while facilitating knowledge sharing across scales.

\subsubsection{Analysis}
\label{sec:analysis}

\begin{figure*}[t]
    \centering
    \includegraphics[width=0.95\linewidth]{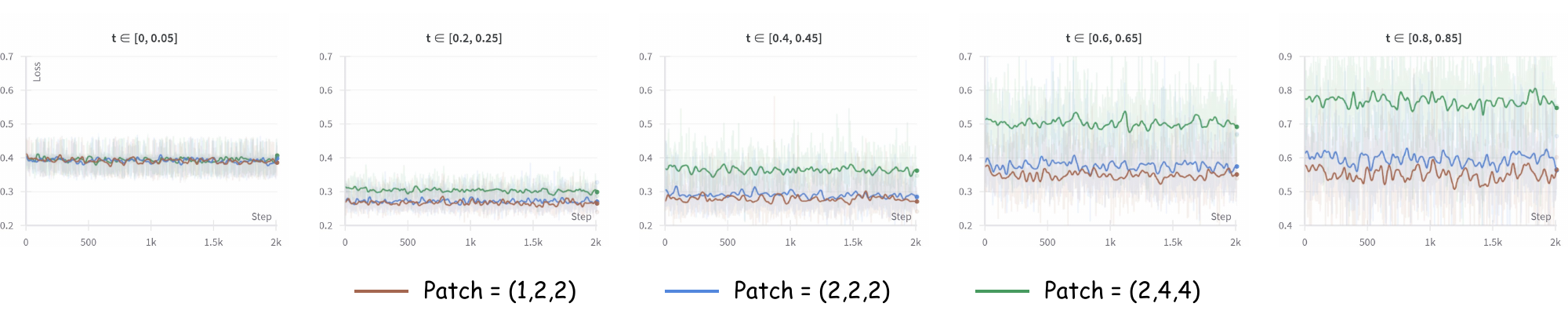}
    \vspace{-1em}
    \caption{Loss curves for different patch sizes at different denoising timesteps. See Sec.~\ref{sec:full-loss-bin} for the complete figure.}
    \label{fig:loss-bin}
\end{figure*}

The unified training with the shared backbone and multiple patchifications provides a novel interface to investigate how different patch sizes affect the quality of denoising prediction, which can be directly reflected by loss magnitudes.

To do so, we evaluate three sets of patch sizes and observe their behavior across timesteps by dividing the denoising process uniformly into 20 time windows. We then analyze the training loss at various timesteps for different patch sizes. The three patch sizes correspond to different scales of spatiotemporal resolution: small, medium, and large.

Figure~\ref{fig:loss-bin} shows the behavior of the loss curves across different patch sizes in 5 uniformly selected $t$ spans, while the complete 20-span visualization is provided in Sec.\ref{sec:full-loss-bin}. In the very early stages of denoising ($0 \leq t \leq 0.1$), the loss curves for all three patch sizes overlap, indicating comparable prediction quality. As $t$ increases, the loss curve for the smallest scale begins to deviate first, posing an obvious and stable gap compared to smaller patches. Similarly, as $t$ grows even larger (say, after $0.4)$, the loss curve for the medium patch also deviates from that of the small patch, implying diverged prediction qualities. These findings validate previous assumptions about the varying nature of tasks performed at different timesteps: early stages of denoising focus on capturing the global structure, where smaller scales are sufficient to predict velocity accurately; as the process progresses, finer details become crucial, necessitating denoising on larger scales. 

This empirical evidence justifies employing a hierarchical generation process by gradually increasing the scale throughout the denoising process, which reduces computational costs while maintaining quality. 

\begin{algorithm}[t]
   \caption{Training of Multi-scale Next-DiT}
   \label{alg:train}
    \begin{algorithmic}
       \STATE {\bfseries Input:} Model $\mathbf{u}^\theta$, dataset $\mathcal{D}$, batch size $B$, patch sizes $\mathcal{P}=\{P_i\}_{i=1}^M$, timeshift values $\{\alpha_i\}_{i=1}^M$, number of training steps $T$, learning rate $\eta$
       \FOR{$iteration=1$ {\bfseries to} $T$}
           \FOR{$k=1$ {\bfseries to} $M$}
               \STATE Sample a batch of $B$ samples from $\mathcal{D}$
               
               \STATE Sample $t'$ uniformly from $[0,1]$
               \STATE Compute rescaled time $t$ with Eq.~\ref{eq:timeshift} and $\alpha_{k}$
               \STATE Compute flow matching loss $\mathcal{L}_k$ with $P_k$
               \STATE Compute gradients $\nabla_{\theta} \mathcal{L}_k$ and accumulate
           \ENDFOR
           \STATE Update model parameters $\mathbf{\theta}$ with $\nabla_{\theta}$
           \STATE Zero out accumulated gradients: $\nabla_{\theta} \mathcal{L} \gets 0$
       \ENDFOR
    \end{algorithmic}
\end{algorithm}

\subsubsection{Training: Scale-aware Timestep Shifting}
\label{sec:scle-aware-ts}
Based on the analysis in Section \ref{sec:analysis}, we observe that larger patch sizes are more suitable for the early stages of denoising with a focus on capturing broad structures. In contrast, smaller patch sizes are more effective in the later stages, where the model needs to capture finer details as the noise diminishes. This observation suggests that applying different timestep sampling schedules to the training of different patch sizes can lead to more efficient training. 

Rather than dividing the trajectory into discrete windows, which limits the interaction between different patch sizes across timesteps and hinders the sharing of knowledge between the various scales, we allow the timestep to be sampled from the entire trajectory $[0,1]$ for all patch sizes. Inspired by Stable Diffusion 3~\citep{sd3}, we then customize the training resources allocation within different scales by assigning different time shift factors.
Specifically, for each patch size $P_i$, we define a shift value $\alpha_i$, which determines how much the sampled timestep is shifted toward the start of the trajectory. 
During training patch size $P_i$, a timestep $t'$ is uniformly sampled from the interval $[0,1]$ for each sample. The timestep $t'$ is then mapped to the actual timestep $t$ using the following formula:
\begin{equation}
    \label{eq:timeshift}
    t = \frac{t'}{t' + \alpha_i - \alpha_i t' }
\end{equation}
For smaller patch sizes, we assign a smaller shift value to increase the likelihood of sampling larger timesteps, (i.e., later stages of denoising), where finer details are more important. 
Conversely, for larger patch sizes, we assign a larger shift value, which increases the probability of sampling smaller timesteps, where coarse structures are more relevant. 

This shift-based allocation effectively tilts the training resources, ensuring that the most advantageous intervals for each patch size, balancing quality and efficiency, are sampled more frequently, thereby maximizing improvements in practical inference.
Furthermore, it allows the model to more effectively share knowledge across stages and patch sizes and learn a unified representation. We summarize the training process in algorithm~\ref{alg:train}.

\begin{figure}[t]
    \centering
    \includegraphics[width=0.8\linewidth]{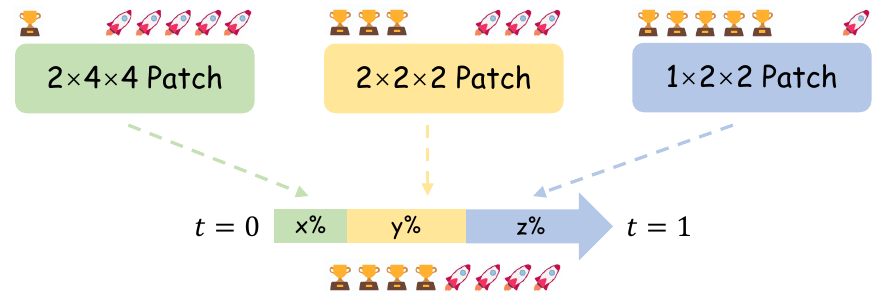}
    \vspace{-0.5em}
    \caption{Multi-scale Patchification allows Lumina-Video to perform flexible multi-stage denoising during inference, leading to a better tradeoff between quality and efficiency.}
    \vspace{-1em}
    \label{fig:inference}
\end{figure}
\subsubsection{Flexible Multi-stage Inference}

Training Multi-scale Next-DiT across multiple scales unlocks significant flexibility during inference. This flexibility is particularly valuable for resource-constrained scenarios, high-throughput requirements, or rapid prototyping when adjusting inference hyperparameters. Following the analysis in Sec.\ref{sec:analysis}, we propose a multi-stage denoising strategy: using smaller scales during early timesteps to determine the video’s general structure, followed by refinement with increasingly larger scales in later stages. As illustrated in Figure \ref{fig:inference}, this approach enjoys the advantage of reducing computation with minor degradation in quality.

\input{table/vbench_extended}

\subsection{Explicit Control over Motion Condition}

Text-to-video models often exhibit overly static behavior, indicating the need for an explicit mechanism for controlling the intensity of motion in the generated videos. In~\sysname{}, we introduce a motion conditioning mechanism that allows for direct control over motion characteristics.

Specifically, we condition the denoising process on a motion score in the same way as it is conditioned on the timestep, shown in Fig.\ref{fig:architecture}(b). 
During training, we calculate this motion score as the average of the magnitude of the optical flow using UniMatch~\citep{unimatch}, an off-the-shelf optical flow model. By conditioning the denoising process on this motion score, the model learns to generate videos with an aligned extent of dynamics. As validated in Sec.~\ref{sec:exp_motion}, by reasonably manipulating the motion conditioning for the positive and negative classifier-free guidance (CFG) samples, the generated dynamic degree can be adjusted effectively and reliably. Moreover, we introduce stochasticity by randomly dropping the motion condition with a probability $p=0.4$ during training to handle situations where the user may not want explicit control over the motion score. This enables the model to adjust the intensity of motion based on the text prompt alone when no motion score is provided.

%% file: table/vbench_extended.tex
\begin{table*}[t]
\centering
\vspace{-0.5em}
\caption{Comparison on VBench. Proprietary models and open-source models are listed separately for better comparison.}
\vspace{-1em}
    \label{tab:vbench}
    \begin{center}
        \begin{small}
            \begin{sc}
                \resizebox{\textwidth}{!}{
\begin{tabular}{lcccccc}
    \toprule
    \textbf{Model} & 
    \textbf{Param} & 
    \makecell{\textbf{Total}\\\textbf{Score} (\%)} & 
    \makecell{\textbf{Quality}\\\textbf{Score} (\%)} & 
    \makecell{\textbf{Semantic}\\\textbf{Score} (\%)} &
    \makecell{\textbf{Motion}\\\textbf{Smoothness} (\%)} &
    \makecell{\textbf{Dynamic}\\\textbf{Degree} (\%)} \\
    \midrule
    \multicolumn{7}{l}{\textit{\textcolor{gray}{Proprietary Models}}} \\[2pt]
    \makecell{Pika-1.0~\citep{pikalabs2024pika}} & 
        - & 
        80.69 &        
        82.92 &        
        71.77 &        
        99.50 &        
        47.50          
        \\
    \makecell{Kling~\citep{kuaishou2024klingai}} & 
        - & 
        81.85 &        
        83.39 &        
        75.68 &        
        99.40 &        
        46.94          
        \\
    \makecell{Vidu~\citep{vidu2025}} & 
        - & 
        81.89 &        
        83.85 &        
        74.04 &        
        97.71 &        
        82.64          
        \\
    \makecell{Gen-3~\citep{gen3alpha2024}} & 
        - & 
        82.32 &        
        84.11 &        
        75.17 &        
        99.23 &        
        60.14          
        \\
    \makecell{Luma~\citep{lumalab2024dreammachine}} & 
        - & 
        83.61 &        
        83.47 &        
        84.17 &        
        99.35 &        
        44.26          
        \\
    \makecell{Sora~\citep{openai2024sora}} & 
        - & 
        84.28 &        
        85.51 &        
        79.35 &        
        98.74 &        
        79.91          
        \\
    \midrule
    \multicolumn{7}{l}{\textit{\textcolor{gray}{Open-Source Models}}} \\[2pt]
    \makecell{OpenSora Plan V1.3~\citep{open-sora-plan}} & 
        2.7B & 
        77.23 &      
        80.14 &      
        65.62 &      
        99.05 &      
        30.28        
        \\
    \makecell{OpenSora V1.2 (8s)~\citep{opensora}} & 
        1.1B & 
        79.76 &      
        81.35 &      
        73.39 &      
        98.50 &      
        42.39        
        \\
    \makecell{VideoCrafter 2.0~\citep{videocrafter2}} & 
        1.4B & 
        80.44 &       
        82.20 &       
        73.42 &       
        97.73 &       
        42.50         
        \\
    \makecell{Allegro~\citep{allegro}} & 
        3B & 
        81.09 &       
        83.12 &       
        72.98 &       
        98.82 &       
        55.00         
        \\
    \makecell{CogVideoX~\citep{CogVideoX}} & 
        5B & 
        81.61 &       
        82.75 &       
        77.04 &       
        96.92 &       
        70.97         
        \\
    \makecell{Pyramid Flow~\citep{PyramidFlow}} & 
        2B &
        81.72 &       
        84.74 &       
        69.62 &       
        99.12 &       
        64.63         
        \\
    \makecell{CogVideoX 1.5~\citep{CogVideoX}} & 
        5B & 
        82.17 &       
        82.78 &       
        79.76 &       
        98.31 &       
        50.93         
        \\
    \makecell{Vchitect 2.0~\citep{fan2025vchitect}} & 
        2B & 
        82.24 &       
        83.54 &       
        77.06 &       
        98.98 &       
        63.89         
        \\
    \makecell{HunyuanVideo~\citep{weijie2024hunyuanvideo}} & 
        13B & 
        83.24 &       
        85.09 &       
        75.82 &       
        98.99 &       
        70.83         
        \\
    \makecell{Lumina-Video (Single Scale)} & 
        2B & 
        82.99 &       
        83.92 &       
        79.27 &       
        98.90 &       
        67.13         
        \\
    \makecell{Lumina-Video (Multi Scale)} & 
        2B & 
        82.94 &       
        84.08 &       
        78.39 &       
        98.92 &       
        71.76         
        \\
    \bottomrule
\end{tabular}
                }
            \end{sc}
        \end{small}
    \end{center}
\end{table*}

%% file: section/4_experiments.tex
\section{Training}
\label{sec:Training}
We adopt a mixture of multiple training strategies:

\textbf{\purple{Progressive Training}} has been widely recognized as an effective and efficient approach for training large-scale visual generative models~\citep{luminamgpt, luminat2x, CogVideoX, open-sora-plan, MovieGen}. Following this paradigm, \sysname{} employs a 4-stage training process, beginning with text-to-image training in the first stage and transitioning to joint text-to-image/video training in the subsequent three stages. Each stage is characterized by a predefined spatial area while allowing variable aspect ratios, ensuring that images and video frames are resized to resolutions close to the specified area while preserving their original aspect ratios. Additionally, each stage is defined with a specific frames-per-second (FPS) value. A maximum clip duration of 4s is applied throughout the entire training.

In \textbf{Stage 1}, the model is trained on pure image data at the resolution of 256 pixels. By rapidly processing a large volume of image-text pairs at high throughput, the model quickly captures the general composition and distribution of visual data, and establishes broad associations between language terms and visual concepts. We use the same image dataset as Lumina-Next~\citep{luminanext}, which shares the similar distribution as JourneyDB~\citep{sun2024journeydb}

In \textbf{Stage 2}, we incorporate the 10M official subset of the large-scale video dataset Panda~\citep{panda}. The spatial resolution remains at 256 pixels, and video frames are extracted at a target FPS of 8. \textbf{Stage 3} raises the spatial resolution to 512 pixels and the FPS to 16, and training is conducted on a mixture of data from OpenVid~\citep{openvid}, Open-Sora-Plan~\citep{open-sora-plan}, and 300k in-house video samples, which consist of diverse real and synthetic data from various sources. Finally, in \textbf{Stage 4}, the spatial resolution increases to 960 pixels, and the FPS is elevated to 24. The same dataset as Stage 3 but filtered with FPS and resolution is used in this stage.

\textbf{\purple{Image-Video Joint Training}} is employed across stages 2 to 4. Leveraging the concept richness and superior quality of image data, this joint training significantly enhances the model’s capacity to understand a broader spectrum of concepts and improves frame-level quality.

\textbf{\purple{Multi-Scale Traning}} is introduced since stage 2. For video, we define three patch sizes: \(P_1 = (1, 2, 2)\), \(P_2 = (2, 2, 2)\), and \(P_3 = (2, 4, 4)\). For image, we use the \(P_1 = (1, 2, 2)\) patchification only. We observe that applying coarser patchifications with temporal compression during image training leads to unintended effects: as images are repeated to fulfill patchification requirements, the model interprets them as \textit{silent videos}, resulting in generated samples with the interesting phenomenon of intra-patch silence and inter-patch dynamics. Scale-aware time-shift introduced in Sec.\ref{sec:scle-aware-ts} is applied throughout progressive training.

\textbf{\purple{Multi-Source Traning}} denotes a novel \textbf{multi-system-prompt training → per-system prompt evaluation → best-subset fine-tuning} strategy. When training relatively large-scale models, the available data typically comes from diverse sources with varying distributions and quality levels. Ideally, we expect the model to learn from all available samples while ensuring that the generated content during inference aligns closely with the quality of the best subset of the training data. However, this objective is challenging, especially when the complexity of data composition makes it difficult for even experienced practitioners to pinpoint which subset qualifies as optimal. 

To solve this problem, we introduce distinct system prompts tailored to each data source into progressive training, and prepend them to the image prompts, forming complete prompts for training. When performing random prompt dropping for CFG, we drop only the image-related prompts while retaining the system prompts to preserve source-specific context during training.

After completing the progressive training, we evaluate the model performance using different system prompts. This evaluation involves both subjective assessments and benchmark-based quantitative metrics. Our observations reveal that generation quality and stability exhibit significant variation across different system prompts.

Based on evaluation results, we conduct a final \textit{best-subset fine-tuning} stage at a reduced learning rate on the most effective system prompt subset. A few hundred iterations suffice to significantly enhance the model’s performance, aligning outputs with the subset’s characteristics while preserving the broad generalization achieved in earlier stages. 

Notably, we find that \textbf{synthetic data}, despite comprising only a small portion ($\sim$10\%) of our in-house dataset, consistently achieves higher scores in per-system prompt evaluation. By checking the generated samples, those from synthetic system prompts also demonstrate greater stability. These findings highlight the effectiveness of synthetic data in video generation, likely due to its simpler distribution, which prevents the model from getting confused by the erratic variability of the real world. To our knowledge, this is the first work to validate synthetic data's utility for large-scale video generation models. We believe these insights will aid research groups with limited resources in developing stronger, more efficient foundational video models.

\input{table/patchification}
\section{Evaluation}

After completing the four stages of progressive training, we select the final checkpoints from stages 3 and 4 and then perform best-subset fine-tuning at the corresponding resolution and frame rate (FPS). This process results in two final models: one with a spatial resolution of \( 512^2 \) at 16 FPS, and another with \( 960^2 \) at 24 FPS. For our quantitative and ablation experiments, we use the first model by default. Demo samples are provided in the supplementary zip file.
\subsection{Comparison with Existing Methods}

\paragraph{Evaluation Benchmark} We quantitatively evaluate the performance of~\sysname{} on VBench~\citep{vbench}, a comprehensive benchmark for text-to-video generation. VBench consists of 16 fine-grained metrics from two primary dimensions including video quality (depicted by quality score) and video-text alignment (depicted by semantic score). During inference, we uniformly sample 70 timesteps and apply a shifting value $\alpha=8.0$. 

We present the quantitative results on VBench in Table~\ref{tab:vbench}, comparing \sysname{} against both proprietary and open-source models. The results demonstrate that \sysname{} is highly competitive overall, performing well in generating high-quality videos while effectively understanding and following user prompts. Detailed results for individual metrics are provided in Table~\ref{tab:vbench_detail} in the appendix.

\subsection{Ablation Study}

\input{table/motion_score}

\subsubsection{Multi-scale Patchification}

\begin{figure}[t]
    \centering
    \includegraphics[width=\linewidth]{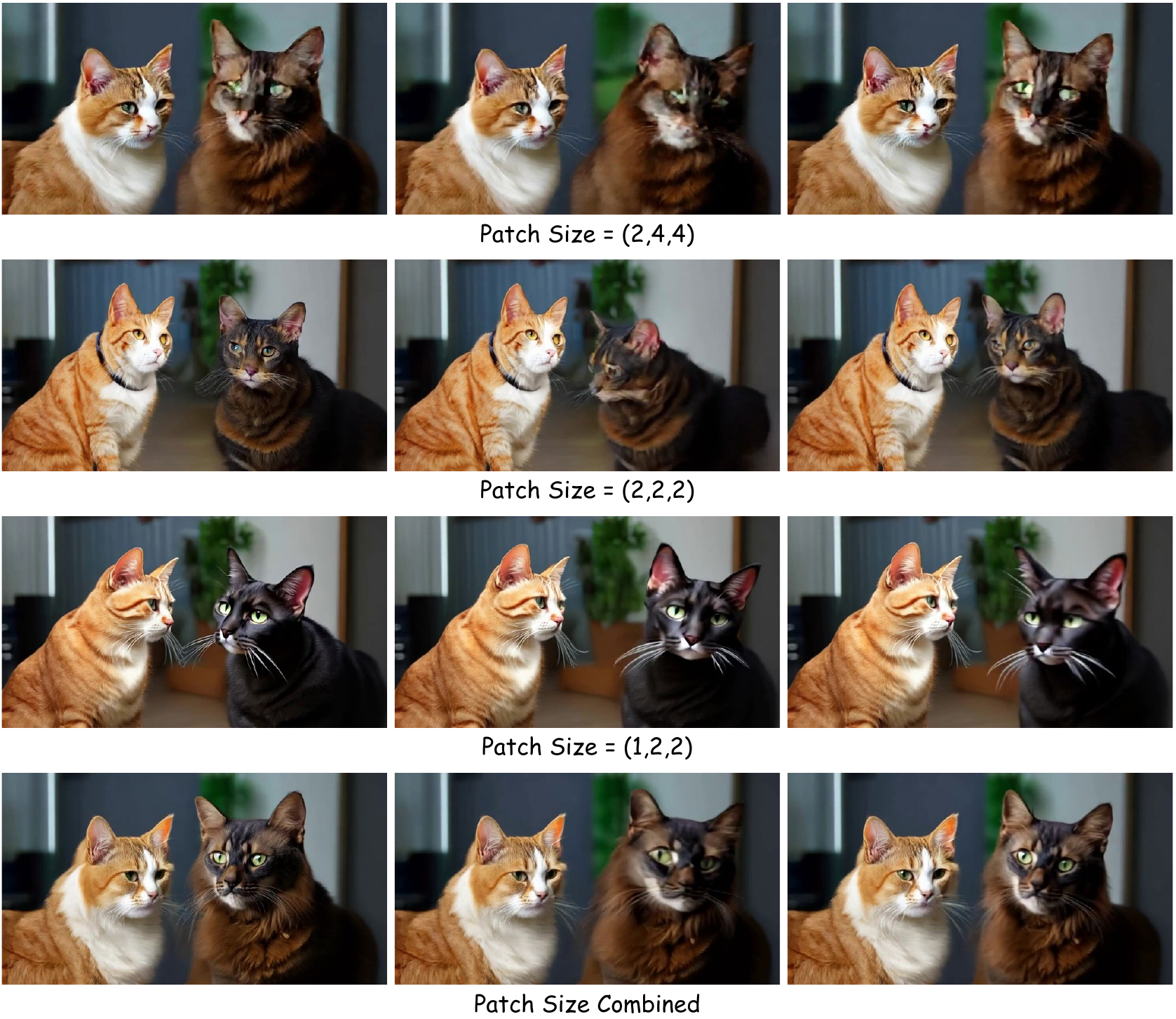}
    \vspace{-2em}
    \caption{Comparison of generated videos using different patchification strategies.}
    \label{fig:patch-size}
    \vspace{-1em}
\end{figure}

Our experimental results in Tab.~\ref{tab:patchification} and Fig.~\ref{fig:patch-size} show that using the smallest patch size throughout the entire process yields the highest overall performance, while larger patch sizes reduce model performance but raise generation speed. Compared to using a single patch size, our patchification stitching-based inference achieves a better efficiency-quality balance. further supporting the relationship between prediction quality, patch size, and timestep illustrated in Fig.~\ref{fig:loss-bin}

\subsubsection{Motion score}
\label{sec:exp_motion}

\begin{figure}[t]
    \centering
    \includegraphics[width=\linewidth]{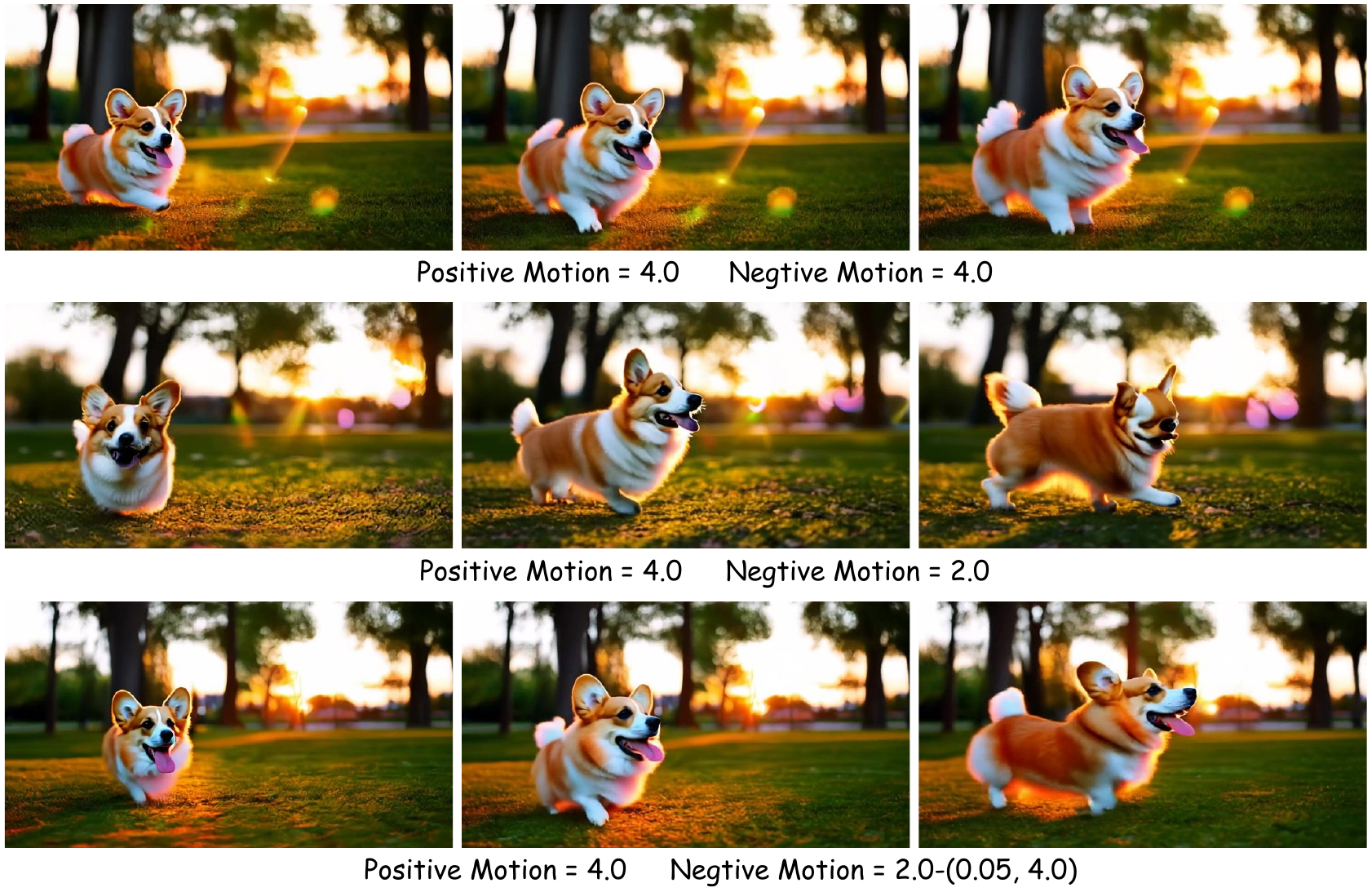}
    \vspace{-2em}
    \caption{Comparison of generated videos using different positive and negative motion scores.}
    \label{fig:motion}
\end{figure}

\sysname{} has introduced motion score as a micro-condition for controlling dynamics. However, as shown in Tab.~\ref{tab:motion_score}, increasing motion conditioning fails to enhance dynamics if the conditioning for CFG negative sample rises simultaneously: setting motion to 4 yields low dynamics, while increasing it to 8 provides only modest improvement.

\textit{This suggests the possibility that high dynamics depends on the difference between positive and negative motion conditioning, not their absolute values}. To verify this, we fixed the negative motion at 2 and observed higher dynamics with [4,2] compared to [4,4], and further improvement with [8,2]. These results confirm motion difference as an effective control for video dynamics.

However, increasing this difference degrades content quality, as reflected in quality and semantic scores. To balance dynamics and quality, we propose initially setting the negative motion to a low value (e.g., 2) and aligning it with positive motion after a threshold (e.g., 0.05). This approach, based on the assumption that motion structures form earlier than fine details in diffusion, achieves an optimal balance, as shown in rows 5–6 in Tab.~\ref{tab:motion_score}. A visualization of motion condition's impact is shown in Fig.~\ref{fig:motion}

%% file: table/patchification.tex
\begin{table}[t]
    \vspace{-0.5em}
    \caption{Ablation study on patchification.}
    \vspace{-1em}
    \label{tab:patchification}
    \begin{center}
        \begin{small}
        \begin{sc}
            \resizebox{\linewidth}{!}{
                \begin{tabular}{ccccc}
                \toprule
                Patch & Time cost & Quality & Semantic & Total
                \\
                \midrule
                (1, 2, 2)            & 1.00       & 83.92         & 79.27          & 82.99       \\
                (2, 2, 2)            & 0.36      & 83.50         & 78.33          & 82.47       \\
                (2, 4, 4)            & 0.07      & 82.47         & 77.77          & 81.53       \\
                Combined       & 0.34       & 84.08         & 78.39          & 82.94      
                \\
                \bottomrule
                \end{tabular}
            }
        \end{sc}
        \end{small}
    \end{center}
\end{table}

%% file: table/motion_score.tex
\begin{table*}[t]
    \vspace{-0.5em}
    \caption{Impact of Motion Score. * means Dynamic Degree is excluded from Quality Score.}
    \label{tab:motion_score}
    \begin{center}
        \begin{small}
            \begin{sc}
                \begin{tabular}{cccccc}
                \toprule
                Positive Motion & Negtive Motion & Dynamic Degree & Quality Score* & Semantic Score & Total Score \\
                \midrule
                4               & 4              & 45.37          & 86.11          & 79.28          & 82.24       \\
                8               & 8              & 53.70          & 86.18          & 78.88          & 82.72       \\
                4               & 2              & 72.78          & 84.11          & 79.10          & 82.41       \\
                8               & 2              & 85.89          & 83.51          & 78.36          & 82.63       \\
                4               & 2-{[}0.05-4{]} & 67.13          & 85.31          & 79.27          & 82.99       \\
                8               & 2-{[}0.05-8{]} & 83.33          & 84.58          & 78.55          & 83.28      \\
                \bottomrule
                \end{tabular}
            \end{sc}
        \end{small}
    \end{center}
    \vspace{-0.5em}
\end{table*}

%% file: section/5_conclusion.tex
\section{Conclusion \& Futer Work}
\label{sec:conclusion}

We present Lumina-Video, a novel framework designed to overcome the unique challenges of video generation by building on the successes of the Next-DiT architecture. Lumina-Video boosts efficiency by utilizing a Multi-scale Next-DiT design and allows direct control of the dynamic degree by explicit conditioning. Through a combination of strategies including progressive training and multi-source training, Lumina-Video demonstrates a powerful solution for generating high-fidelity videos with both spatial and temporal consistency. Additionally, the companion Lumina-V2A model further enhances real-world applicability through audio-visual synchronization.

Moving forward, we will focus on two key areas. First, multi-scale patchification shares the same motivation as dynamic neural networks, where simpler tasks require fewer resources. Insights from dynamic networks and network compression suggest that an organic compression across multiple dimensions (e.g., depth, width, tokens) generally offers better trade-offs than focusing on a single dimension, and we will explore this further. Second, while benchmarks show that Lumina-Video generates prompt-coherent videos meeting basic quality standards, with higher standards, gaps remain compared to commercial solutions in video aesthetics, complex motion synthesis, and artifact-free details. These challenges will drive our efforts in data curation, architecture refinement, and pipeline optimization.

\newpage
\section{Impact Statements}
This paper discusses a video generation method. Video generation, as a promising technology, also comes with significant societal risks, many of which are common to generative models in general. These risks deserve careful consideration and detailed discussion. However, the potential impacts of our method are reflective of those of the broader video generation field. Therefore, we believe there is no need to specifically highlight any particular impact in this context.

%% file: section/supp.tex
\section{Basic Compositions of~\sysname{}}
\label{sec:basic_comp}
\subsection{Loss function}
\sysname{} is trained with flow matching \cite{FlowMatching,FlowMatchingGuide,RectifiedFlow}, a generative framework which builds a probability path from random noise $\mathbf{x}_0 \sim \mathcal{N}(\mathbf{0}, \mathbf{I})$ to the target data distribution $\mathbf{x}_1 \sim p_{\mathrm{data}}$. 
The model is tasked with predicting the velocity field of samples $\mathbf{u}_t(\cdot)$ which can be used to reconstruct the data sample by solving the following Ordinary Differential Equation (ODE):
\begin{equation}
    \frac{\mathrm{d}}{\mathrm{d}t} \mathbf{x}_t = \mathbf{u}_t (\mathbf{x}_t)
\end{equation}
One commonly adopted form of the probability path is the linear path, which assumes $\mathbf{x}_t$ is a linear interpolation of noise and data sample:
\begin{equation}
\label{eq:xt}
    \mathbf{x}_t = t \mathbf{x}_1 + (1-t) \mathbf{x}_0 \sim \mathcal{N}(t \mathbf{x}_1, (1-t)^2 \mathbf{I})
\end{equation}
This assumption leads to a velocity field in the form of
\begin{equation}
\label{eq:velocity}
    \mathbf{u}_t(\mathbf{x}|\mathbf{x}_1) = \frac{\mathbf{x}_1 - \mathbf{x}}{1 - t}
\end{equation}
Equipped with Equation \ref{eq:xt} and \ref{eq:velocity}, the flow matching loss can be formulated as
\begin{equation}
    \mathcal{L}(\theta) = \mathbb{E}_{t,\mathbf{x}_t,\mathbf{x}_1} \lVert \mathbf{u}^\theta_t(\mathbf{x_t},t) - \mathbf{u}_t(\mathbf{x}_t|\mathbf{x}_1) \rVert^2
\end{equation}

\subsection{Architecture}

\paragraph{VAE.} We adopt the 3D causal VAE of CogVideoX \cite{CogVideoX} for encoding and decoding videos between the pixel space and the latent space. Compared to the 2-D VAE used in Lumina-Next with spatial compression only, this VAE achieves higher efficiency by applying compression in the temporal dimension with less information leak from redundant frames. 

\paragraph{Text Encoder.} 
Following Lumina-Next, we utilize the Gemma-2-2B model \cite{team2024gemma} as our text encoder. Despite being a lightweight text encoder, Gemma-2-2B excels at extracting visual semantics from natural language, enabling accurate and efficient alignment between textual descriptions and video content. 

\paragraph{Multi-scale Next-DiT.} The backbone of Lumina-Video is an improved version of Next-DiT~\citep{luminanext}, a flow-based diffusion transformer that incorporates the following key modifications to diffusion transformers: 
1) Replacing 1D RoPE with 3D RoPE to instill more accurate positional prior in visual modeling;
2) Introducing sandwich normalization to control the magnitude of activations and stabilize the training process;
3) Incorporating Grouped-Query Attention to reduce computational demand;
To adapt Next-DiT to video generation, Lumina-Video introduces Multi-scale Next-DiT, a transformative extension of Next-DiT to a multi-scale architecture, which is elaborated in Sec.~\ref{sec:ms-next-dit}.

\section{Training Details}
The AdamW optimizer~\citep{adamw}, with weight decay set to \(0.0\) and betas \((0.9, 0.95)\), is employed. Furthermore, PyTorch FSDP~\citep{fsdp} with gradient checkpointing is utilized for reduced memory cost. To enhance training throughput, all video data are pre-encoded using a VAE encoder before training. The data are then clustered based on duration, ensuring that each global batch consists of samples with similar lengths.

\newpage
\section{Detailed Results}
\subsection{Full VBench Results}
\input{table/vbench_all}
\newpage
\subsection{Full version of Figure~\ref{fig:loss-bin}}
\label{sec:full-loss-bin}
\begin{figure*}[h]
    \centering
    \includegraphics[width=0.95\linewidth]{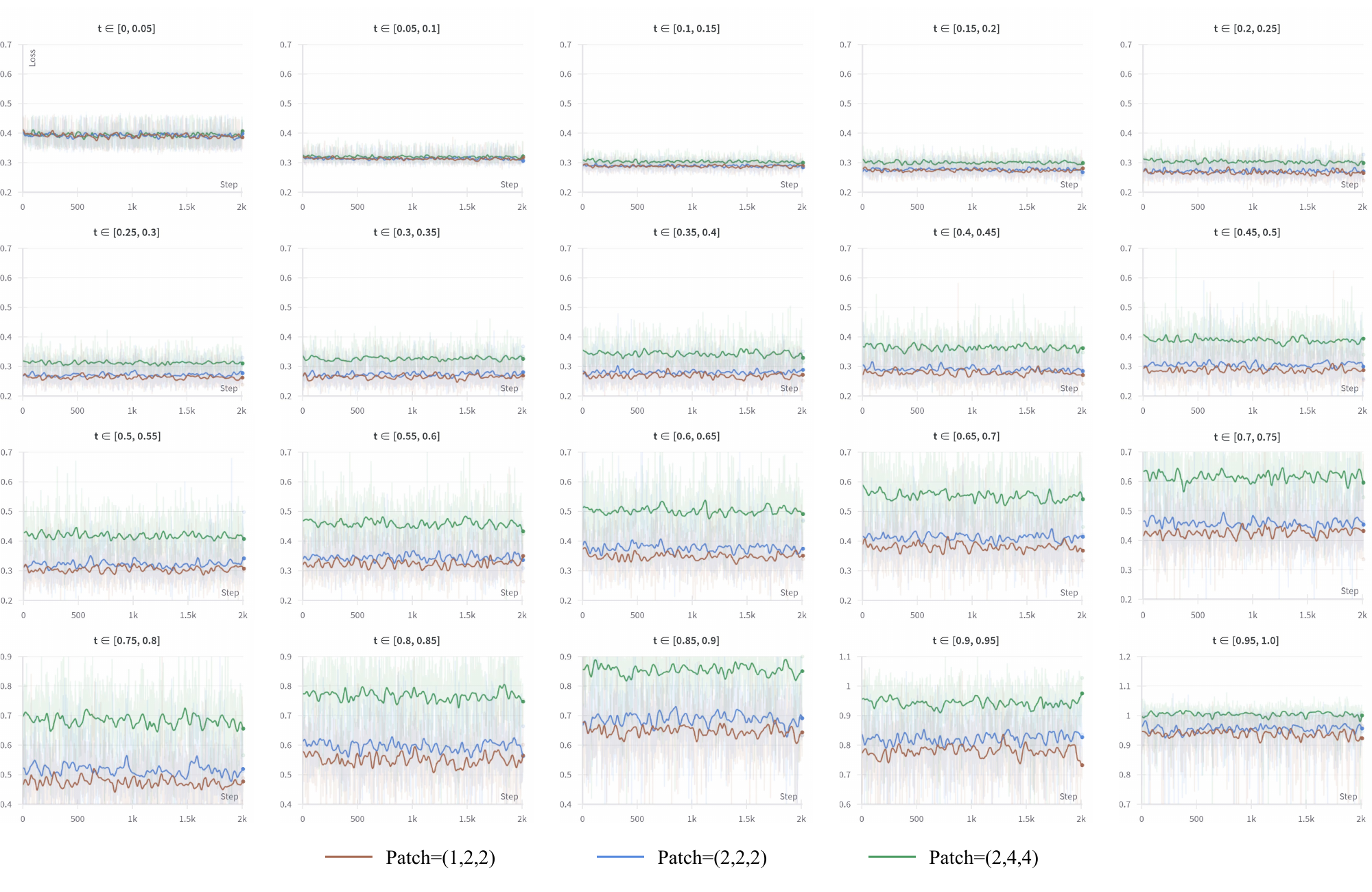}
    \vspace{-1.5em}
    \caption{Complete figure of loss curves for different patch sizes at different denoising timesteps.}
    \vspace{-1em}
    \label{fig:loss-bin-complete}
\end{figure*}

\section{Video-to-Audio}
\label{sec:V2A}
In this section, we extend Lumina-Video with video-to-audio ability by designing Lumina-V2A to generate ambient sounds for silent video by synchronizing with visible scenes.
\subsection{Background}
Some existing works adopt a two-stage process to first align the video features with acoustic~\cite{luo2024diff, wangfrieren} by unsupervised pretraining, and then introduce diffusion or flow-matching models to generate audio. Other approaches are proposed to first extract visual language~\cite{wang2024v2a} or time-varying features~\cite{zhang2024foleycrafter, jeong2024read, lee2024video, xie2024sonicvisionlm} such as timestamps or energy curves from videos and then leverage pre-trained text-to-audio generation models to produce corresponding audio via trainable introduced adapters. Recent V2A works have achieved audio generation conditioned on both video and text~\cite{polyak2024movie, cheng2024taming, chen2024video}, creating high-fidelity sound effects aligned with visual content. Our proposed V2A model aims to generate audios that are temporally synchronized with videos and semantically aligned with both video and text.          

\begin{figure}[t]
    \centering
    \includegraphics[width=1\linewidth]{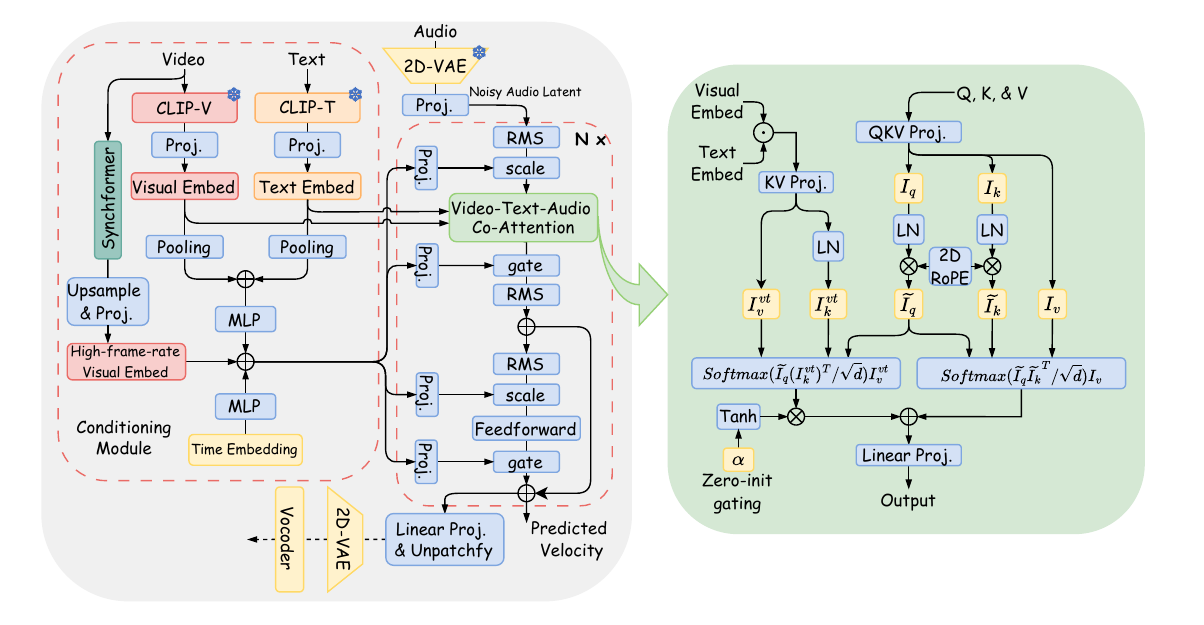}
    \vspace{-2em}
    \caption{Illustration of Lumina-V2A Model based on Next-DiT}
    \label{fig:v2a_architecture}
    \vspace{-1em}
\end{figure}

\subsection{Model Architecture}
As depicted in Figure~\ref{fig:v2a_architecture}, our Lumina-V2A model receives the video and text conditions to generate audio based on Next-DiT and rectified flow matching~\cite{RectifiedFlow}. Specifically, the input audio waveform is first transformed into the 2D mel-spectrogram with an STFT operator and is then encoded by a pre-trained audio VAE encoder~\cite{liu2023audioldm} to obtain the compressed audio latents. Meanwhile, the pre-trained CLIP visual encoder and CLIP textual encoder~\cite{fang2023data, radford2021learning} are employed to successively encode video and text into frame-level visual features and text embeddings. Next, visual features, text embeddings and audio latents are independently projected into modality-specific inputs to undergo the following Next-DiT blocks. 

To efficiently integrate video, text, and audio modalities, we adopt a sequence of video-text-audio Next-DiT blocks to process the concatenated multimodal features via an inner co-attention mechanism. Within the co-attention module, the visual and text embeddings are concatenated together to interact with audio latent tokens like Lumina-Video. Additionally, It is important to ensure semantic alignment between audio and video-text and temporal synchronization between audio and video. To do so, we propose a multimodal conditioning module to integrate the time embedding, global visual and textual features, and high-frame-rate visual features from Synchformer~\cite{iashin2024synchformer}, forming a multimodal condition to be injected into Next-DiT blocks via scaling and gating operation. 

During the inference stage, once the audio latent representation is generated by the proposed diffusion transformer, the pre-trained VAE decoder is used to reconstruct generated audio latents back to the mel-spectrogram that is then transformed into audio waveform via a pre-trained HiFi-GAN vocoder~\cite{kong2020hifi}. 

\subsection{Training Data}
We conduct experiments of our video-to-audio model on the VGGSound~\cite{chen2020vggsound}, a large-scale audio-visual dataset including 500 hours of videos with audio tracks in the wild and 310 classes. After filtering out invalid video IDs, we split the remaining dataset into around 180k videos for training, 2k videos for validation, and 15k for testing. To further improve the quality of VGGSound, we adopt the AV-Align score~\cite{yariv2024diverse} as the temporal alignment metric to select more aligned audio-visual pairs by setting the threshold as 0.2, resulting in about 110k high-quality video-audio pairs for future fine-tuning. By following the existing work~\cite{cheng2024taming}, we truncate each video clip to 8s duration during the training stage.

%% file: table/vbench_all.tex
\begin{table}[htbp]
    \centering
    \caption{Detailed results on VBench. -ss means single scale inference and -ms means multi-scale inference}
    \label{tab:vbench_detail}
    \resizebox{\textwidth}{!}{
    \begin{tabular}{lcccccccc}
        \toprule
        Model & 
        \makecell{Subject\\Consistency} & 
        \makecell{Background\\Consistency} & 
        \makecell{Temporal\\Flickering} &
        \makecell{Motion\\Smoothness} &
        \makecell{Dynamic\\Degree} &
        \makecell{Aesthetic\\Quality} &
        \makecell{Imaging\\Quality} &
        \makecell{Object\\Class} \\
        \midrule
        \multicolumn{9}{l}{\textit{\textcolor{gray}{Proprietary Models}}} \\

        Pika-1.0      & 96.94 & 97.36 & 99.74 & 99.50 & 47.50 & 62.04 & 61.87 & 88.72 \\
        Kling         & 98.33 & 97.60 & 99.30 & 99.40 & 46.94 & 61.21 & 65.62 & 87.24 \\
        Vidu          & 94.63 & 96.55 & 99.08 & 97.71 & 82.64 & 60.87 & 63.32 & 88.43 \\
        Gen-3 Alpha   & 97.10 & 96.62 & 98.61 & 99.23 & 60.14 & 63.34 & 66.82 & 87.81 \\
        Luma          & 97.33 & 97.43 & 98.64 & 99.35 & 44.26 & 65.51 & 66.55 & 94.95 \\
        Sora          & 96.23 & 96.35 & 98.87 & 98.74 & 79.91 & 63.46 & 68.28 & 93.93 \\
        \midrule
        \multicolumn{9}{l}{\textcolor{gray}{Open-Source Models}} \\
        OpenSora Plan V1.3 & 97.79 & 97.24 & 99.20 & 99.05 & 30.28 & 60.42 & 56.21 & 85.56 \\
        OpenSora V1.2 & 96.75 & 97.61 & 99.53 & 98.50 & 42.39 & 56.85 & 63.34 & 82.22 \\
        VideoCrafter 2.0 & 96.85 & 98.22 & 98.41 & 97.73 & 42.50 & 63.13 & 67.22 & 92.55 \\
        Allegro       & 96.33 & 96.74 & 99.00 & 98.82 & 55.00 & 63.74 & 63.60 & 87.51 \\
        CogVideoX     & 96.23 & 96.52 & 98.66 & 96.92 & 70.97 & 61.98 & 62.90 & 85.23 \\
        Pyramid Flow   & 96.95 & 98.06 & 99.49 & 99.12 & 64.63 & 63.26 & 65.01 &86.67 \\    
        CogVideoX 1.5 & 96.87 & 97.35 & 98.88 & 98.31 & 50.93 & 62.79 & 65.02 & 87.47 \\
        Vchitect 2.0  & 96.83 & 96.66 & 98.57 & 98.98 & 63.89 & 60.41 & 65.35 & 86.61 \\
        HunyuanVideo  & 97.37 & 97.76 & 99.44 & 98.99 & 70.83 & 60.36 & 67.56 & 86.10 \\
        Lumina-Video-ss &96.06	&97.26	&98.63	&98.90	&67.13	&62.27	&64.58	&91.03\\
        Lumina-Video-ms & 95.95	&96.99	&98.59	&98.92	&71.76	&62.25	&63.85	&90.69\\
       \midrule
       \midrule
        Model & 
        \makecell{Multiple\\Objects} & 
        \makecell{Human\\Action} & 
        \makecell{Color} & 
        \makecell{Spatial\\Relationship} & 
        \makecell{Scene} & 
        \makecell{Appearance\\Style} & 
        \makecell{Temporal\\Style} & 
        \makecell{Overall\\Consistency} \\
        \midrule
        \multicolumn{9}{l}{\textit{\textcolor{gray}{Proprietary Models}}} \\

        Pika-1.0      & 43.08 & 86.20 & 90.57 & 61.03 & 49.83 & 22.26 & 24.22 & 25.94 \\
        Kling         & 68.05 & 93.40 & 89.90 & 73.03 & 50.86 & 19.62 & 24.17 & 26.42 \\   
        Vidu          & 61.68 & 97.40 & 83.24 & 66.18 & 46.07 & 21.54 & 23.79 & 26.47 \\
        Gen-3 Alpha   & 53.64 & 96.40 & 80.90 & 65.09 & 54.57 & 24.31 & 24.71 & 26.69 \\
        Luma          & 82.63 & 96.40 & 92.33 & 83.67 & 58.98 & 24.66 & 26.29 & 28.13 \\
        Sora          & 70.85 & 98.20 & 80.11 & 74.29 & 56.95 & 24.76 & 25.01 & 26.26 \\
        \midrule
        \multicolumn{9}{l}{\textcolor{gray}{Open-Source Models}} \\
        OpenSora Plan V1.3 & 43.58 & 86.80 & 79.30 & 51.61 & 36.73 & 20.03 & 22.47 & 24.47 \\
        OpenSora V1.2 & 51.83 & 91.20 & 90.08 & 68.56 & 42.44 & 23.95 & 24.54 & 26.85 \\
        VideoCrafter 2.0 & 40.66 & 95.00 & 92.92 & 35.86 & 55.29 & 25.13 & 25.84 & 28.23 \\
        Allegro       & 59.92 & 91.40 & 82.77 & 67.15 & 46.72 & 20.53 & 24.23 & 26.36 \\
        CogVideoX     & 62.11 & 99.40 & 82.81 & 66.35 & 53.20 & 24.91 & 25.38 & 27.59 \\
        Pyramid Flow   & 50.71 & 85.60 & 82.87 & 59.53 & 43.20 & 20.91 & 23.09 & 26.23 \\
        CogVideoX 1.5 & 69.65 & 97.20 & 87.55 & 80.25 & 52.91 & 24.89 & 25.19 & 27.30 \\
        Vchitect 2.0  & 68.84 & 97.20 & 87.04 & 57.55 & 56.57 & 23.73 & 25.01 & 27.57 \\
        HunyuanVideo  & 68.55 & 94.40 & 91.60 & 68.68 & 53.88 & 19.80 & 23.89 & 26.44 \\
        Lumina-Video-ss & 68.32	&97.67	&90.16	&67.27	&56.08	&23.64	&25.66	&28.22 \\
        Lumina-Video-ms &65.27	&97.33	&89.58	&63.68	&57.10	&25.47	&23.42	&28.23 \\
        \bottomrule
    \end{tabular}
    }
\end{table}